\documentclass[journal]{IEEEtran}

\usepackage{amsmath,amsfonts,amssymb}
\usepackage{bm}
\usepackage[varg]{newtxmath}

\usepackage{algorithm}
\usepackage{algpseudocode}

\usepackage{array}
\usepackage{booktabs}
\usepackage{multirow}
\usepackage{makecell}
\usepackage{threeparttable}
\usepackage{adjustbox}
\usepackage[table]{xcolor}

\usepackage{graphicx}
\usepackage{subcaption}
\usepackage{textcomp}
\usepackage{pifont}
\usepackage{ragged2e}
\usepackage{verbatim}
\usepackage{url}
\usepackage{cite}

\usepackage{hyperref}
\usepackage{orcidlink}

\newcommand{\cmark}{\ding{51}}

\newcommand{\pp}{\,\mathrm{pp}}

\usepackage{tabularx}
\usepackage{placeins}
\newcommand{\BestScore}[1]{\textbf{#1}}
\newcommand{\SecondScore}[1]{\underline{#1}}

\hypersetup{
    pdftitle={GeoJeV: Unified Spatial Probability for Training-Free Efficient Evidence Acquisition in Ultra-High-Resolution Remote Sensing},
    pdfauthor={Lingfeng Chen, Xinyi Niu, Panhe Hu, Shuanghui Zhang, Li Liu, Zhen Liu, and Yongxiang Liu}
}

\begin{document}

% GeoJeV journal manuscript; layout retained from the supplied bi-graph template.
% Build: latexmk -pdf -interaction=nonstopmode -file-line-error -halt-on-error -outdir=out GeoJeV.tex
\title{GeoJeV: Unified Spatial Probability for Training-Free Efficient Evidence Acquisition in Ultra-High-Resolution Remote Sensing}

\author{
        Xinyi Niu$^{\orcidlink{0009-0003-5777-9852}}$,
        Lingfeng Chen$^{\orcidlink{0009-0003-2690-9407}}$,
        Jiehua Zhang$^{\orcidlink{0000-0003-4342-0139}}$,  
        Panhe Hu$^{\orcidlink{0000-0001-5895-3152}}$, 
        Li Liu$^{\orcidlink{0000-0002-2011-2873}}$,
        Zhen Liu$^{\orcidlink{0000-0002-1233-1494}}$,
        and~Yongxiang Liu$^{\orcidlink{0000-0002-0682-8365}}$

\thanks{This work was supported in part by Fundamental and Interdisciplinary Disciplines Breakthrough Plan of the Ministry of Education of China under Grant JYB2025XDXM110, in part by National Natural Science Foundation of China under Grant 62401585, in part by the Science and Technology Innovation Program of Hunan Province under Grant 2024RC3120.}
\thanks{Xinyi Niu, Lingfeng Chen, Jiehua Zhang, Panhe Hu, Li Liu, Zhen Liu and Yongxiang Liu are with the College of Electronic Science and Technology, National University of Defense Technology, Changsha 410073, Hunan, China. E-mail: (niuxinyi23@nudt.edu.cn; chenlingfeng@nudt.edu.cn;\ jiehua.zhang@oulu.fi;
hupanhe13@nudt.edu.cn; \nolinkurl{liuli_nudt@nudt.edu.cn};\
zhen\_liu@nudt.edu.cn; lyx\_bible@sina.com).
 \textit{(Xinyi Niu and Lingfeng Chen contributed equally to this work.)(Corresponding authors: Panhe Hu and Zhen Liu.)}}
}

% The paper headers
\markboth{Journal of \LaTeX\ Class Files,~Vol.~14, No.~8, August~2021}%
{Niu \MakeLowercase{\textit{et~al.}}: GeoJeV: Unified Spatial Probability for Training-Free Efficient Evidence Acquisition in Ultra-High-Resolution Remote Sensing}

\maketitle

% ===== Begin inlined source: sec/0_abstract.tex =====

\begin{abstract}
Ultra-high-resolution (UHR) remote sensing imagery provides rich spatial details but poses challenges for accurate and efficient multimodal understanding. Existing evidence acquisition methods selectively explore relevant image regions, yet largely rely on generative multimodal large language models (MLLMs) for intermediate search control. Reinforcement learning from human feedback (RLHF) primarily aligns these models with human-preferred responses rather than explicitly optimizing region-level relevance judgments, while repeated autoregressive decoding introduces additional computational overhead.
To address these limitations, we propose \textbf{GeoJeV}, a training-free method that formulates evidence acquisition as spatial probability-driven decision making. At each evaluated region, a frozen JeV model produces a probability distribution over candidate regions.
Specifically, Region Rank prioritizes the candidates, Sparse Branch adaptively sets the branching width through sparse probability support, and Depth Gate controls refinement depth through branch sparsity and distribution concentration. By coordinating search priority, branching width, and refinement depth through a shared distribution, GeoJeV reduces reliance on RLHF-aligned generation and eliminates intermediate decoding. Experiments on LRS-GRO, MME-RealWorld-RS, and LRS-VQA achieve accuracies of 60.59\%, 52.79\%, and 34.32\%, respectively. Across the three benchmarks, GeoJeV improves the average accuracy of Qwen3-VL-8B by $+5.32\pp$ with only $1.12$ additional seconds per question, while achieving a $7.8\times$ inference speedup over the recent method. Code will be available at \href{https://github.com/Niuxinyi02/GeoJeV}{\texttt{https://github.com/Niuxinyi02/GeoJeV}}.
\end{abstract}


% ===== End inlined source: sec/0_abstract.tex =====
% ===== Begin inlined source: sec/1_intro.tex =====
% Integrated from evid_intro_related.tex.
\section{Introduction}
\label{sec:intro}

\IEEEPARstart{U}{ltra-high-resolution} (UHR) remote sensing imagery has been
widely used in urban planning, disaster monitoring, and environmental perception~\cite{zhu2017deep,lobry2020rsvqa,xlrsbench2025,guo2024saan,lin2025hslabeling,zhang2026rpun}. Its high spatial resolution provides rich spatial details, making it possible to recognize small objects and local structures that are difficult to distinguish in lower-resolution imagery~\cite{han2021small,xia2018dota,zhang2025orsisam,zhan2025hbipidinet}.
However, UHR images usually contain a large number of pixels, which leads
to high computational cost for multimodal understanding~\cite{geollava8k2025,zhang2025mmerealworld}. Processing the whole image at its original resolution is expensive, while direct downsampling can lead to the loss of important visual details, making small objects difficult to recognize~\cite{xlrsbench2025,luo2025textpruning}. Therefore, a key challenge
in UHR remote sensing understanding is how to preserve task-relevant visual
information while reducing unnecessary computation over large-scale scenes.

\begin{figure*}[t]
    \centering
    \includegraphics[width=0.95\linewidth]{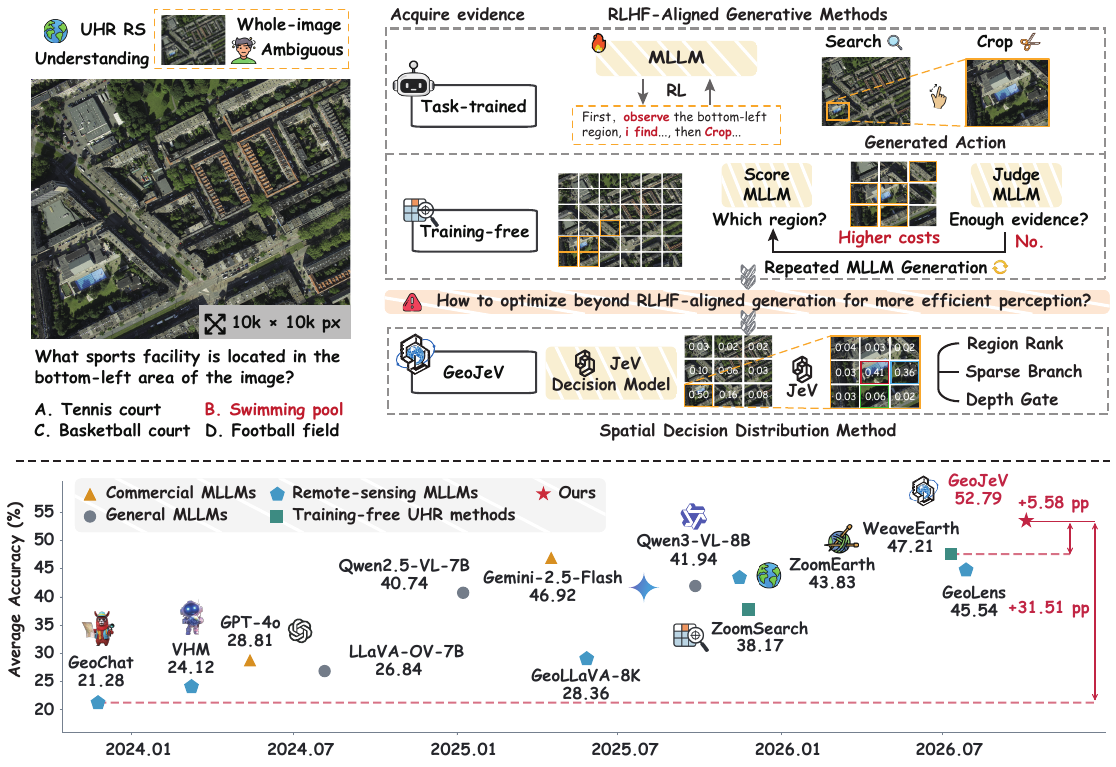}
    \caption{Motivation and performance of GeoJeV. (Top) Evidence-acquisition pipeline: a shared spatial probability distribution jointly controls search priority, branching width, and refinement depth. (Bottom) Average accuracy on MME-RealWorld-RS, where GeoJeV with Qwen3-VL-8B leads all plotted methods, exceeding WeaveEarth by $+5.58\pp$.}
\label{fig:motivation}
\end{figure*}

Existing studies mainly address this challenge in two ways. One line
of work improves the visual representation of UHR images by dividing original
images into smaller patches or processing them at multiple scales,
thereby preserving fine-grained details and spatial
context~\cite{guo2024llavauhd,li2024monkey,zhang2026llavauhdv2,geollava8k2025,li2025lsknet}.
Another line of research focuses on selective evidence acquisition through
progressive localization and analysis of task-relevant
regions~\cite{wu2024vstar,li2025dyfo,zhang2026geovis},
thereby concentrating computation on informative
regions~\cite{zoomsearch2025,weaveearth2026}.
This selective allocation is particularly suitable for UHR remote sensing
scenes, where task-relevant objects often occupy only a small part
of the image.

However, existing evidence acquisition methods still largely rely on generative multimodal large language models (MLLMs) to control the intermediate search process. Training-based methods learn to generate search or cropping actions through reinforcement
learning (RL)~\cite{liu2026zoomearth,geoeyes2026,geolens2026,zheng2026deepeyes}, while training-free methods repeatedly query MLLMs to score candidate regions or assess evidence
sufficiency~\cite{wu2024vstar,shen2025zoomeye,gazeearth2026}.
These intermediate decisions require accurately judging the
relevance of local visual evidence to the
task~\cite{pang2025vhm,gao2025damrs}.
Yet reinforcement learning from human feedback (RLHF) primarily
aligns these models with human-preferred responses rather than
explicitly optimizing region-level relevance
judgments~\cite{ouyang2022training}.
Moreover, repeatedly generating intermediate decisions through
autoregressive decoding introduces additional computational overhead.
As shown in Fig.~\ref{fig:motivation}, these limitations motivate
a natural question:
\emph{can evidence acquisition move beyond RLHF-aligned generation
for more efficient perception?}

The JeV model addresses this through probabilistic decision making. Trained with Reinforcement Learning for Calibrated Decisions (RLCD), JeV directly optimizes probabilistic decisions by assigning probability distributions to candidate options~\cite{typesafe2026jev}. This capability motivates a probability-based approach to evaluating candidate regions in UHR images, where the relevance of each region can be estimated based on the question and available visual evidence. Such probability estimates can guide evidence acquisition while avoiding repeated autoregressive decoding for intermediate judgments.

Building on this, we introduce \textbf{GeoJeV}, a training-free
method for efficient UHR remote sensing understanding. GeoJeV formulates evidence acquisition as spatial probability-driven decision making.
Given a question and the current visual observation, the JeV model
produces a probability distribution over candidate regions at each
evaluated node. Region Rank prioritizes candidate regions, Sparse Branch adaptively determines branching width, and Depth Gate controls refinement depth through early stopping.
By coordinating these three aspects through a shared distribution,
GeoJeV grounds search control in region-level relevance probabilities,
reducing its reliance on RLHF-aligned generation and
eliminating repeated autoregressive decoding of intermediate decisions.
The acquired regions are combined into a spatial evidence canvas for final answering by a frozen MLLM.

Our main contributions are summarized as follows:
\begin{itemize}

    \item We propose GeoJeV, a training-free method for efficient UHR remote sensing understanding that formulates evidence acquisition as spatial probability-driven decision making, moving beyond RLHF-aligned generation and eliminating intermediate decoding.

    \item We develop a unified spatial probability distribution that jointly governs search priority, branching width, and refinement depth via Region Rank, Sparse Branch, and Depth Gate, enabling adaptive evidence acquisition while reducing redundant exploration.

    \item We conduct extensive experiments on LRS-GRO, MME-RealWorld-RS, and LRS-VQA to systematically evaluate overall accuracy, computational efficiency, and cross-backbone generalization, complemented by component ablation and parameter sensitivity analyses.

\end{itemize}

% \begin{itemize}

%     \item We propose GeoJeV, a training-free method that reformulates evidence acquisition for UHR remote sensing understanding as spatial probability-driven decision making. A frozen JeV model guides visual search without intermediate language generation, while a separate frozen MLLM performs final answering.

%     \item We design a unified acquisition mechanism in which Region Rank, Sparse Branch, and Depth Gate exploit the order, sparse support, and geometry of a shared spatial probability distribution to control search priority, branching width, and refinement depth. Early stopping reuses cached parent information to retain child crops while skipping their JeV evaluation and further subdivision.

%     \item Experiments on LRS-GRO, MME-RealWorld-RS, and LRS-VQA demonstrate improved average accuracy across two frozen answering backbones and a favorable accuracy--latency trade-off. Component ablations on MME-RealWorld-RS further support the efficiency benefits of adaptive branching and early stopping.

% \end{itemize}

\section{Related Work}
\label{sec:related_work}

This study focuses on training-free efficient evidence acquisition based on spatial probability distributions for UHR remote sensing
understanding. This section reviews efficient UHR perception, adaptive
evidence acquisition, and preference alignment and probabilistic decisions. 

\subsection{Efficient UHR Remote Sensing Perception}
\label{sec:related_efficiency}

\noindent\textbf{Resolution-Adaptive Visual Encoding.}
UHR imagery understanding requires preserving fine spatial details while controlling the number of visual tokens. Recent work on MLLMs has therefore
explored flexible high-resolution image encoding. LLaVA-NeXT adopts a
multi-view strategy to process images at different resolutions, while
LLaVA-OneVision further improves visual representation across diverse
image scales~\cite{liu2024llavanext,li2024llavaonevision}.
LLaVA-UHD introduces adaptive image partitioning, token compression, and
spatial organization to support images with varying resolutions and aspect
ratios~\cite{guo2024llavauhd}.
Encoding local regions at higher resolution reduces the loss
of fine detail caused by global downsampling. Preserving these details is
particularly important for UHR remote sensing imagery, where small objects
and local structures often occupy only a small fraction of the
image~\cite{han2021small,zhang2025mmerealworld}.
Resolution-adaptive encoding thus provides richer visual information for
local object recognition and scene understanding.

\noindent\textbf{Visual Token Selection and Compression.}
Another line of work reduces the cost of UHR understanding by removing
redundant visual information or allocating computation more
selectively~\cite{chen2024fastv,yang2025visionzip,zhang2025sparsevlm,shang2025prumerge}.
These methods select informative tokens or merge similar features to
form compact visual representations.
GeoLLaVA-8K~\cite{geollava8k2025} introduces background token pruning and
anchored token selection to retain informative content while supporting
remote sensing images up to 8K resolution.
Text-guided token pruning has also been explored for UHR remote sensing,
where relevant image tiles are progressively selected through a dynamic
image pyramid and redundant visual tokens are removed from coarse to
fine~\cite{luo2025textpruning}.
These methods improve inference efficiency by reducing visual redundancy,
making UHR image understanding more practical under limited computational
resources.

\subsection{Adaptive Evidence Acquisition}
\label{sec:related_acquisition}

\noindent\textbf{Training-Based Visual Exploration.}
Training-based methods learn to locate and examine task-relevant image
regions through supervised fine-tuning (SFT) or
RL~\cite{shao2024visualcot,su2025pixelreasoner}.
V$^{*}$~\cite{wu2024vstar} introduces language-guided visual search with
a specially trained visual search model to locate detailed information.
ZoomEarth~\cite{liu2026zoomearth} learns cropping and zooming behaviors
with region-guided rewards, while GeoEyes~\cite{geoeyes2026} uses
evidence gain as its reward during zoom interactions.
GeoLens~\cite{geolens2026} learns multi-tool visual reasoning, enabling
the model to select and invoke visual operations according to the current
reasoning state.
These methods demonstrate the effectiveness of allocating computation
according to task demands, but generally require additional training to
learn specialized acquisition behaviors.

\noindent\textbf{Training-Free Evidence Acquisition.}
Training-free methods use frozen models to acquire task-relevant visual
evidence during inference.
Visual Sketchpad~\cite{hu2024visualsketchpad} uses tools such as cropping
and zooming to inspect image content during reasoning.
ZoomEye~\cite{shen2025zoomeye} organizes image regions into a hierarchy for
progressive inspection, while ZoomSearch~\cite{zoomsearch2025} explores multiple
branches and arranges selected patches in their original spatial layout.
For evidence construction, Retrieval-Augmented Perception~\cite{wang2025rap}
retrieves task-relevant crops and adjusts the amount of visual evidence
using retrieval scores and model confidence.
WeaveEarth~\cite{weaveearth2026} selects spatially complementary regions
and combines them with global context to form compact visual evidence.
These methods support reuse across datasets and answering backbones
without task-specific parameter updates.
However, methods that rely on repeated language generation for region
selection or stopping decisions can incur additional inference costs.

\subsection{Preference Alignment and Probabilistic Decisions}
\label{sec:related_decisions}

\noindent\textbf{Preference-Aligned Generation.}
Preference alignment guides MLLMs toward human-preferred responses after
SFT.
RLHF optimizes model responses using rewards learned from human
comparisons~\cite{ouyang2022training}, while Direct Preference Optimization
(DPO) learns directly from preference pairs without an explicit reward
model~\cite{rafailov2023direct}.
Preference learning may also favor excessive agreement with users or
longer responses~\cite{sharma2024sycophancy,singhal2024length}.
Alongside response alignment, VHM~\cite{pang2025vhm} emphasizes visual
grounding through SFT with factual and deceptive questions.
During visual search, an MLLM must compare candidate regions and
determine whether further inspection is likely to provide useful
information for answering the current question.
Response-level preference alignment improves instruction following and
response quality, but does not directly optimize region-level relevance
judgments for evidence acquisition.

\noindent\textbf{Probabilistic Decision Modeling.}
Probabilistic decision modeling focuses on comparing candidate options
through explicit probability estimates.
JeV~\cite{typesafe2026jev} is trained with RLCD to optimize calibrated
decisions over candidate options rather than human-preferred responses.
This objective aims to make its confidence estimates reflect decision
reliability.
JeV directly returns structured probabilities without autoregressive text
generation or response parsing, reducing the overhead of repeated
decisions.
Metask-JEV~\cite{metaskjev2026} provides an openly available decision
model that returns candidate probabilities in a single forward pass.
AnyJev~\cite{zhang2026anyjev} instead reads option probabilities from
existing instruction-tuned models without further training and corrects
biases caused by option labels and ordering.
These probability outputs support direct comparison and ranking of
candidate options without intermediate language generation, providing
a useful basis for evidence selection and search control.

\noindent\textbf{Our Insights.}
These studies highlight the need for training-free evidence acquisition
that balances accuracy and computational efficiency.
GeoJeV addresses these needs through spatial probability-driven decision making.
At each evaluated node, a frozen JeV model estimates a probability
distribution over candidate regions.
Region Rank, Sparse Branch, and Depth Gate use this shared distribution
to control search priority, branching width, and refinement depth.
This reduces reliance on RLHF-aligned generation for search control and
avoids repeated decoding of intermediate decisions.
The selected evidence is organized into a spatial evidence canvas for final answering by a frozen MLLM, without task-specific parameter updates.

\section{Methodology}
\label{sec:method}

\begin{figure*}[t]
\centering
\includegraphics[width=0.95\linewidth]{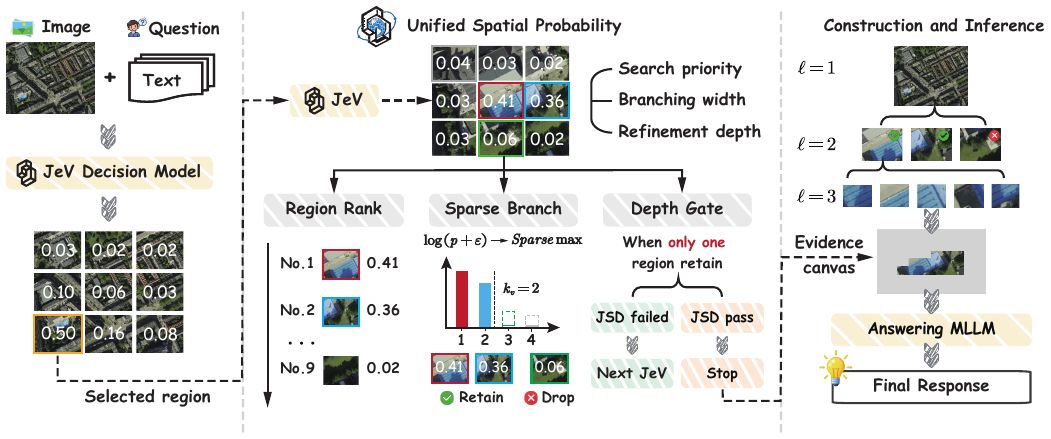}
\caption{Overview of GeoJeV. (Left) JeV estimates a spatial probability distribution over candidate regions. (Middle) Region Rank sets search priority, Sparse Branch adapts branching width, and Depth Gate controls refinement depth via early stopping. (Right) The acquired regions are combined into a spatial evidence canvas for final answering.}
\label{fig:method}
\end{figure*}

GeoJeV formulates evidence acquisition for UHR remote sensing
understanding as spatial probability-driven decision making.
This training-free formulation uses a frozen JeV model for
search control without intermediate autoregressive decoding,
while a separate frozen MLLM performs final answering.
Following the framework in Fig.~\ref{fig:method}, we present
the problem formulation, spatial probability distribution,
adaptive search, and spatial evidence construction.

\subsection{Overview and Problem Formulation}
\label{subsec:overview}

Given a UHR image $I\in\mathbb R^{H\times W\times3}$
and a question $q$, GeoJeV
uses a JeV model $f_\phi$ for search control and an MLLM
$a_\theta$ for final answering, with frozen parameters $\phi$
and $\theta$, respectively. Search and evidence selection produce
the final evidence set $\mathcal F^\star$. The function $C$
combines the selected regions into a spatial evidence canvas
$\widetilde I$. The answering MLLM then uses this canvas to
predict $\widehat y$:
\vspace{0.3em}
\begin{equation}
\label{eq:overview}
\widetilde I=C(I,\mathcal F^\star),\qquad
\widehat y=a_\theta(\widetilde I,q).
\end{equation}

The original full image serves as the root of the search tree.
Each node $v$ has a bounding box $B_v$, a crop
$I_v=\operatorname{Crop}(I,B_v)$, and a depth $\ell_v$.
The root $v_0$ has depth one. Dividing a node into a
$g\times g$ grid produces $n=g^2$ child boxes $B_{v,i}$,
where $g\ge2$ is an integer and $i\in\{1,\ldots,n\}$.
The set $\mathcal J_v$ contains the indices of children whose
width and height are both at least $\delta>0$ pixels.
The number of valid children is $m_v=|\mathcal J_v|$.

For $\min(H,W)\ge\delta$, the maximum search depth is
$L=1+\lfloor\log_g(\min(H,W)/\delta)\rfloor$, held fixed
within each benchmark.
A node is terminal if $\ell_v\ge L$ or $m_v=0$.
Positive integers $K$ and $M$ cap the number of retained children
per expansion and the number of JeV forward passes per question,
respectively. The same $K$ also limits the final evidence count.
Root evaluation counts toward $M$, while terminal and early-stop
checks do not. Section~\ref{sec:experiment_setup} provides the
numerical settings.

\subsection{Spatial Probability Distribution}
\label{subsec:spatial_distribution}

For each node $v$ to be expanded, we add a grid on $I_v$
before JeV preprocessing to obtain the visual input $V_v$.
Given $V_v$ and the question $q$, JeV computes logits
$\mathbf z_v=f_\phi(V_v,q)\in\mathbb R^n$ for all $n$
candidate regions in a single forward pass.
We then apply softmax with temperature $T$ to obtain
\begin{equation}
\label{eq:spatial_probability}
p_{v,i}=
\frac{\exp(z_{v,i}/T)}
{\sum_{j=1}^{n}\exp(z_{v,j}/T)},
\qquad
\mathbf p_v\in\Delta^{n-1},
\end{equation}
where $T>0$ is the temperature and $z_{v,i}$ is the logit
for candidate region $i$. The probabilities in
$\mathbf p_v=(p_{v,1},\ldots,p_{v,n})$ sum to one over all
$n$ candidate regions. Only children with indices in
$\mathcal J_v$ can be expanded. These probabilities describe
the relative relevance of each region to the question and
guide the search.

\subsection{Probability-Guided Adaptive Search}
\label{subsec:adaptive_search}

\subsubsection{Order-based Region Rank}
\label{subsubsec:region_rank}

Region Rank orders the valid candidate regions by decreasing
$p_{v,i}$. Let $\pi_v=(\pi_{v,1},\ldots,\pi_{v,m_v})$
denote this ordering, with $\pi_{v,j}$ the index at rank $j$.
Candidate region $i$ receives the search priority
\begin{equation}
\label{eq:region_priority}
c_{v,i}=
\frac{p_{v,i}-p_v^-}
{p_v^+-p_v^-+\epsilon_r},
\qquad i\in\mathcal J_v,
\end{equation}
where $p_v^-$ and $p_v^+$ are the minimum and maximum
probabilities over $\mathcal J_v$, and $\epsilon_r>0$
prevents division by zero. Each selected child $u_i$ is
assigned the search priority $c_{u_i}=c_{v,i}$.
We use $c_v$ to denote the search priority of node $v$ and
set $c_{v_0}=1$ for the root. This normalization preserves
the probability ordering among children of the same node.
It also places their priorities on a common scale for
comparison across nodes in the search queue and during
final evidence selection.

\subsubsection{Support-based Sparse Branch}
\label{subsubsec:sparse_branch}

Sparse Branch adapts the branching width through sparse
probability support. We first transform valid probabilities
into log scores $x_{v,i}=\log(p_{v,i}+\epsilon_s)$, where
$\epsilon_s>0$ is a stabilizer. For the log-score vector
$\mathbf x_v=(x_{v,i})_{i\in\mathcal J_v}$,
Sparsemax~\cite{martins2016sparsemax} gives
\begin{equation}
\label{eq:sparse_branch_projection}
\mathbf w_v=
\underset{\mathbf w\in\Delta^{m_v-1}}{\arg\min}
\frac12\lVert\mathbf w-\mathbf x_v\rVert_2^2,
\end{equation}
where $\Delta^{m_v-1}$ is the simplex over the $m_v$
valid candidate regions. Let $\mathbf w_v=(w_{v,i})_{i\in\mathcal J_v}$ denote the output
weight vector, with $w_{v,i}$ representing the weight of
candidate region $i$.
Sparsemax can assign zero weights to
some candidates. Its positive support is defined as
$\mathcal S_v=\{i\in\mathcal J_v:w_{v,i}>0\}$.
The branching width and retained candidate set are then given by
\begin{equation}
\label{eq:sparse_branch_selection}
k_v=\max\{1,\min(K,|\mathcal S_v|)\},\qquad
\mathcal R_v=\{\pi_{v,1},\ldots,\pi_{v,k_v}\},
\end{equation}
where $k_v$ is the number of retained branches and
$\mathcal R_v$ contains the retained indices. The positive support $\mathcal S_v$ identifies the highest ranked
candidates, and $\mathcal R_v$ retains up to $K$ of them in
probability order. Comparable scores can
support multiple branches, while larger score gaps favor
a smaller branching width.

\subsubsection{Geometry-based Depth Gate}
\label{subsubsec:depth_gate}

Depth Gate uses branching width and probability concentration
to control early stopping. We sort $\mathbf p_v$ from largest
to smallest to obtain $\overline{\mathbf p}_v$. We then compare
it with the concentrated distribution $\mathbf e_n=(1,0,\ldots,0)$
and the uniform distribution $\mathbf u_n=(1/n,\ldots,1/n)$:
\begin{equation}
\label{eq:gate_margin}
\rho_v=
D_{\mathrm{JS}}(\overline{\mathbf p}_v,\mathbf u_n)
-
D_{\mathrm{JS}}(\overline{\mathbf p}_v,\mathbf e_n),
\end{equation}
where $D_{\mathrm{JS}}$ is the Jensen--Shannon (JS)
divergence~\cite{lin1991divergence}, computed using base-two
logarithms. When $\rho_v\ge0$, the distribution is at least
as close to $\mathbf e_n$ as to $\mathbf u_n$.
Sorting makes the comparison independent of region positions.
The gate $\zeta_v$ at node $v$ and the early-stop flag $s_u$
for a retained child $u$ are defined as
\begin{equation}
\label{eq:depth_gate}
\begin{aligned}
\zeta_v
&=\mathbb I[m_v=n\land k_v=1\land\rho_v\ge0],\\
s_u
&=\mathbb I[\ell_u\in\mathcal D_{\mathrm{gate}}]\zeta_v,
\end{aligned}
\end{equation}
where $\mathbb I$ is the indicator function and
$\mathcal D_{\mathrm{gate}}\subseteq\{2,\ldots,L-1\}$
specifies the depths at which early stopping is allowed.
The condition $m_v=n$ requires all candidate regions to meet
the minimum size $\delta$.

\subsubsection{Search Execution}
\label{subsubsec:search_execution}

Algorithm~\ref{alg:geojev} initializes a search tree $\mathcal T$
and a priority queue $\mathcal O$ with the original full image
as the root. The counter $N_{\mathrm{dec}}$ records the number
of JeV forward passes. Each expansion uses one forward pass
to compute the spatial probability distribution shared by
Region Rank, Sparse Branch, and Depth Gate.
Children selected by $\mathcal R_v$ are created in rank order.
Successfully created children are added to $\mathcal T$ under
the current node and placed in $\mathcal O$ with their assigned
priorities and early-stop flags. Their insertion order in
$\mathcal T$ is recorded. Nodes with equal priorities follow
this order in both the queue and final evidence selection.

The search terminates when $\mathcal O$ is empty or
$N_{\mathrm{dec}}=M$. The final leaf set $\mathcal L(\mathcal T)$
forms the candidate evidence set, including unexpanded nodes
remaining in $\mathcal O$. From this set, at most $K$ regions
are retained following the selection procedure in
Section~\ref{subsec:evidence}.
For $L\ge2$, the number of JeV forward passes satisfies
\begin{equation}
\label{eq:search_complexity}
N_{\mathrm{dec}}
\le
\min\left\{
M,\sum_{d=1}^{L-1}K^{d-1}
\right\},
\end{equation}
where $d$ denotes search depth. This bound applies only to
JeV forward passes and excludes the computational costs of
candidate generation and final answering.

\begin{algorithm}[t]
\caption{Inference Procedure of GeoJeV}
\label{alg:geojev}
\small
\begin{algorithmic}[1]
\Require Image $I$, question $q$, frozen models $f_\phi,a_\theta$
\Ensure Answer $\widehat y$

\Statex \textbf{Initialize}
\State Create full-image root $v_0$ with
$\ell_{v_0}=1$, $c_{v_0}=1$, and $s_{v_0}=0$
\State $\mathcal T\gets\{v_0\}$; $\mathcal O\gets\{v_0\}$;
$N_{\mathrm{dec}}\gets0$

\Statex \textbf{Search}
\While{$\mathcal O\ne\varnothing$ \textbf{and} $N_{\mathrm{dec}}<M$}
    \State Pop the highest-priority node $v$ from $\mathcal O$, breaking ties by insertion order
    \If{$\ell_v\ge L$ \textbf{or} $s_v=1$}
        \State \textbf{continue}
    \EndIf
    \State Subdivide $B_v$ into $g\times g$ candidate boxes $B_{v,i}$
    \State Obtain valid indices $\mathcal J_v$ using the size limit $\delta$
    \If{$\mathcal J_v=\varnothing$}
        \State \textbf{continue}
    \EndIf
    \State Obtain $\mathbf p_v$ with one JeV forward pass and softmax
    \State $N_{\mathrm{dec}}\gets N_{\mathrm{dec}}+1$
    \State Use \textsc{RegionRank} to obtain $\pi_v$ and
    $c_{v,i}$ for all $i\in\mathcal J_v$
    \State $(k_v,\mathcal R_v)\gets
    \Call{SparseBranch}{\mathbf p_v,\mathcal J_v,\pi_v}$
    \State $\zeta_v\gets
    \Call{DepthGate}{\mathbf p_v,\mathcal J_v,k_v}$
    \For{$i\in\mathcal R_v$ in rank order}
        \State Construct child $u_i$ from $B_{v,i}$
        \If{construction fails}
            \State \textbf{continue}
        \EndIf
        \State $B_{u_i}\gets B_{v,i}$;
        $\ell_{u_i}\gets\ell_v+1$; $c_{u_i}\gets c_{v,i}$
        \State $s_{u_i}\gets
        \mathbb I[\ell_{u_i}\in\mathcal D_{\mathrm{gate}}]\,\zeta_v$
        \State Attach $u_i$ to $v$ in $\mathcal T$ and record its insertion order
        \State Insert $u_i$ into $\mathcal O$ with priority $c_{u_i}$
    \EndFor
\EndWhile

\Statex \textbf{Select evidence and answer}
\State $\mathcal F\gets\mathcal L(\mathcal T)$
\State Sort $\mathcal F$ by decreasing $c_v$, using insertion order for ties
\State Retain the first $\min(K,|\mathcal F|)$ candidates as $\mathcal F^\star$
\State $\widetilde I\gets C(I,\mathcal F^\star)$
\State \Return $a_\theta(\widetilde I,q)$
\end{algorithmic}
\end{algorithm}

\subsection{Evidence Construction and Final Answering}
\label{subsec:evidence}

Evidence construction selects regions from the final leaf set
$\mathcal F=\mathcal L(\mathcal T)$ and combines them into a
spatial evidence canvas for final answering. The full image
serves as the canvas when the root is the only candidate.

\subsubsection{Capacity-limited Selection}
Final evidence selection ranks the leaf candidates by their
inherited search priorities $c_v$ and retains at most $K$ regions.
The selected evidence set is
\begin{equation}
\label{eq:evidence_consolidation}
\mathcal F^\star=
\{v_{(j)}:1\le j\le\min(K,|\mathcal F|)\},
\end{equation}
where $v_{(j)}$ is the candidate at rank $j$ after sorting by
decreasing $c_v$. Candidates with equal priorities follow
their insertion order in $\mathcal T$, consistent with the
ordering used during search.

\subsubsection{Spatial Composition and Final Answering}
For each selected region $v\in\mathcal F^\star$, the function
$C$ extracts a crop from the original image $I$ using its
bounding box $B_v$. The crops are arranged according to their
locations in the original image and combined into the spatial
evidence canvas $\widetilde I=C(I,\mathcal F^\star)$.
The canvas presents the selected local details together with
their relative spatial positions. The frozen answering MLLM
then takes $\widetilde I$ and the question $q$ as input and
produces the final answer $\widehat y=a_\theta(\widetilde I,q)$.

% ===== End inlined source: sec/3_method.tex =====
% ===== Begin inlined source: sec/4_experiment.tex =====
\section{Experiments}
\setlength{\floatsep}{5pt}
\setlength{\textfloatsep}{5pt}
\label{sec:experiment}

\subsection{Experimental Setup}
\label{sec:experiment_setup}

We evaluate GeoJeV in terms of overall and category-wise
accuracy, the accuracy--latency trade-off, and generalization
across frozen answering backbones. Component ablations examine
the contributions of Region Rank, Sparse Branch, and Depth
Gate to answering accuracy and evidence acquisition cost, and
sensitivity analyses vary the branching width, grid
granularity, and temperature.

\begin{figure}[!t]
\centering
\includegraphics[width=0.9\columnwidth]{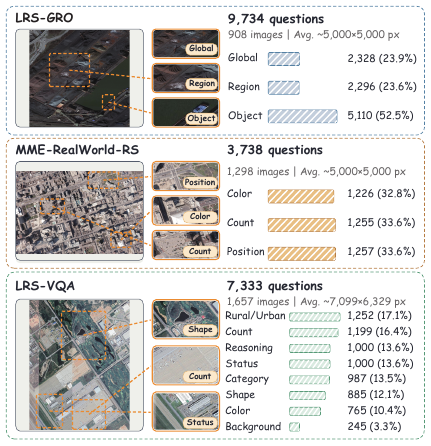}
\caption{Overview of the three UHR benchmarks, including example images, image counts, average resolutions, and category-wise question distributions.}
\label{fig:dataset_statistics}
\end{figure}

\noindent \textbf{Evaluation Datasets.}
We evaluate GeoJeV on the three UHR remote sensing benchmarks, namely LRS-GRO~\cite{liu2026zoomearth}, MME-RealWorld-RS~\cite{zhang2025mmerealworld}, and LRS-VQA~\cite{luo2025textpruning}. Fig.~\ref{fig:dataset_statistics} illustrates their image resolutions and question distributions.

\textit{i) LRS-GRO.}
LRS-GRO contains 908 images and 9,734 questions across 17 question types, grouped into Global, Region, and Object levels. With an average image resolution of approximately $5{,}000\times5{,}000$ pixels, it evaluates visual understanding at different spatial scales, from global scene comprehension to local object perception.

\textit{ii) MME-RealWorld-RS.}
MME-RealWorld-RS comprises 1,298 high-resolution images and 3,738 manually annotated questions across the Color, Count, and Position categories. Its average image resolution exceeds $5{,}000\times5{,}000$ pixels. These tasks emphasize fine-grained visual perception and spatial localization, making the benchmark suitable for evaluating evidence acquisition.

\textit{iii) LRS-VQA.}
LRS-VQA consists of 1,657 images and 7,333 question--answer pairs spanning the eight categories of Rural/Urban, Count, Reasoning, Status, Category, Shape, Color, and Background. With an average resolution of $7{,}099\times6{,}329$ pixels, it provides a comprehensive evaluation of visual perception and reasoning across diverse remote sensing tasks.

% ===== Begin inlined source: figures/dataset_statistics.tex =====
% ===== End inlined source: figures/dataset_statistics.tex =====

\noindent \textbf{Baseline Methods.}
We compare GeoJeV with representative methods from four categories.

\textit{i) Commercial MLLMs.}
We include GPT-4o~\cite{openai2024gpt4o} and Gemini-2.5-Flash~\cite{geminiteam2025} as commercial MLLM baselines.

\textit{ii) General-purpose MLLMs.}
We include Qwen2.5-VL-7B~\cite{bai2025qwen25vl}, Qwen3-VL-8B~\cite{bai2025qwen3vl}, LLaVA-v1.6-7B~\cite{liu2024llavanext}, and LLaVA-OV-7B~\cite{li2024llavaonevision}. These general-purpose MLLMs provide different capabilities for high-resolution visual understanding.

\textit{iii) Remote sensing MLLMs.}
We compare with GeoChat~\cite{kuckreja2024geochat}, VHM~\cite{pang2025vhm}, GeoLLaVA-8K~\cite{geollava8k2025}, GeoLens~\cite{geolens2026}, and ZoomEarth~\cite{liu2026zoomearth}, covering remote sensing understanding, high-resolution perception, and adaptive evidence acquisition.

\textit{iv) Training-free evidence acquisition methods.}
We include ZoomSearch~\cite{zoomsearch2025} and WeaveEarth~\cite{weaveearth2026}. These methods use frozen models to acquire task-relevant visual evidence through selective region exploration and spatial evidence construction, without task-specific parameter updates.

We compare Direct, ZoomSearch, WeaveEarth, and GeoJeV using Qwen3-VL-8B as the shared frozen answering backbone. To evaluate generalization across backbones, we use two frozen answering MLLMs with the same JeV model and search settings. Then we conduct ablation studies to evaluate the contribution of each search control component.

\noindent \textbf{Evaluation Metrics and Protocol.}
We report accuracy, end-to-end inference latency, and the average number of JeV forward passes per question. Benchmark accuracy is the unweighted mean of category-wise accuracies, with three categories for LRS-GRO and MME-RealWorld-RS and eight for LRS-VQA. Latency is measured in seconds per question and includes evidence acquisition, tool execution, spatial evidence canvas construction, and final answering. The forward-pass count includes root evaluations when applicable and excludes final answering, as counted in Algorithm~\ref{alg:geojev}.

We use the official test splits and evaluation scripts for all three benchmarks. Generated answers are evaluated against the reference answers using the same criteria for all methods within each benchmark. All component ablations are conducted in the same execution environment.

% ===== Begin inlined source: tables/main_lrs_mme.tex =====
\begin{table*}[!t]
\centering
\caption{Category-wise accuracy comparison on LRS-GRO and MME-RealWorld-RS.}
\label{tab:main_lrs_mme}
\small
\setlength{\tabcolsep}{3pt}
\renewcommand{\arraystretch}{1.0}
\begin{threeparttable}
\begin{tabular*}{\textwidth}{@{\extracolsep{\fill}}l*{8}{c}@{}}
\toprule
\multirow{2}{*}{\textbf{Method}} &
\multicolumn{4}{c}{\textbf{LRS-GRO} (\%) $\uparrow$} &
\multicolumn{4}{c}{\textbf{MME-RealWorld-RS} (\%) $\uparrow$} \\
\cmidrule(lr){2-5}\cmidrule(lr){6-9}
& Global & Region & Object & Avg.
& Color & Count & Position & Avg. \\
\midrule
\rowcolor[gray]{0.94}
\multicolumn{9}{@{}l}{\textit{Commercial MLLMs}} \\
GPT-4o~\cite{openai2024gpt4o} & 31.10 & 12.37 & 21.86 & 21.78 & 34.18 & 15.17 & 37.07 & 28.81 \\
Gemini-2.5-Flash~\cite{geminiteam2025} & 57.95 & 23.39 & 30.76 & 37.37 & 55.86 & \BestScore{33.28} & 51.63 & 46.92 \\
\midrule
\rowcolor[gray]{0.94}
\multicolumn{9}{@{}l}{\textit{General-purpose MLLMs}} \\
Qwen2.5-VL-7B~\cite{bai2025qwen25vl} & 69.39 & 38.47 & 43.58 & 50.48 & 45.94 & 21.05 & 55.23 & 40.74 \\
Qwen3-VL-8B~\cite{bai2025qwen3vl} & \SecondScore{76.25} & 42.03 & 54.77 & 57.68 & 48.07 & 21.35 & 56.41 & 41.94 \\
LLaVA-v1.6-7B~\cite{liu2024llavanext} & 75.52 & 35.67 & 45.24 & 52.14 & 24.51 & 18.03 & 27.49 & 23.34 \\
LLaVA-OV-7B~\cite{li2024llavaonevision} & 62.43 & 36.18 & 39.92 & 46.18 & 26.14 & 27.57 & 26.81 & 26.84 \\
\midrule
\rowcolor[gray]{0.94}
\multicolumn{9}{@{}l}{\textit{Remote sensing MLLMs}} \\
% RSUniVLM & 63.02 & 34.10 & 37.55 & 44.89 & 28.97 & 19.00 & 28.33 & 25.43 \\
GeoChat~\cite{kuckreja2024geochat} & 62.43 & 36.79 & 40.72 & 46.65 & 23.11 & 15.66 & 25.06 & 21.28 \\
VHM~\cite{pang2025vhm} & 60.40 & 26.92 & 35.03 & 40.78 & 20.32 & 16.80 & 35.24 & 24.12 \\
GeoLLaVA-8K~\cite{geollava8k2025} & 10.44 & 11.96 & 11.19 & 11.20 & 27.92 & 22.27 & 34.90 & 28.36 \\
GeoLens~\cite{geolens2026} & \BestScore{79.00} & \SecondScore{48.90} & \SecondScore{57.60} & \BestScore{61.83} & 53.23 & 19.74 & \SecondScore{63.64} & 45.54 \\
ZoomEarth~\cite{liu2026zoomearth} & 67.45 & 42.31 & 51.96 & 53.91 & 49.69 & 31.66 & 50.14 & 43.83 \\
\midrule
\rowcolor[gray]{0.94}
\multicolumn{9}{@{}l}{\textit{Training-free evidence acquisition methods}} \\
ZoomSearch~\cite{zoomsearch2025} & 67.14 & 46.78 & 52.04 & 55.32 & 41.67 & 17.86 & 54.97 & 38.17 \\
WeaveEarth~\cite{weaveearth2026} & 75.34 & 47.74 & 56.50 & 59.86 & 51.39 & 27.65 & 62.60 & 47.21 \\
% \midrule
% \multicolumn{9}{@{}l}{\textit{General focus methods for MLLMs}} \\
% AttWarp & 77.75 & 49.00 & 53.39 & 60.05 & 54.58 & 19.00 & 62.13 & 45.24 \\
% ViCrop & 77.45 & 51.00 & 53.82 & 60.76 & 51.87 & 17.54 & 62.13 & 43.85 \\
\midrule
\rowcolor[gray]{0.94}
\multicolumn{9}{@{}l}{\textit{Ours}} \\
\textbf{GeoJeV (Qwen3-VL-8B)} & 73.02 & \BestScore{51.05} & \BestScore{57.69} & \SecondScore{60.59} & \BestScore{69.80} & 29.93 & 58.63 & \SecondScore{52.79} \\
\textbf{GeoJeV (LLaVA-OV-7B)} & 69.80 & 46.65 & 49.75 & 55.40 & \SecondScore{68.69} & \SecondScore{31.81} & \BestScore{67.14} & \BestScore{55.88} \\
\bottomrule
\end{tabular*}
\par\vspace{2pt}
\begin{tablenotes}[flushleft]
\footnotesize\raggedright
\item[] \textbf{Avg.}: Unweighted mean across categories. Best and second-best results are in \textbf{bold} and \underline{underlined}, respectively.
\end{tablenotes}
\end{threeparttable}
% ===== End inlined source: tables/main_lrs_mme.tex =====

% Keep the two main-result tables together when preceding text changes.
\par\vspace{\dblfloatsep}

% ===== Begin inlined source: tables/main_lrsvqa.tex =====
\normalsize
\caption{Category-wise accuracy comparison on LRS-VQA.}
\label{tab:main_lrsvqa}
\small
\setlength{\tabcolsep}{3pt}
\renewcommand{\arraystretch}{1.0}
\begin{threeparttable}
\begin{tabular*}{\textwidth}{@{\extracolsep{\fill}}l*{9}{c}@{}}
\toprule
\multirow{2}{*}{\textbf{Method}} &
\multicolumn{9}{c}{\textbf{LRS-VQA} (\%) $\uparrow$} \\
\cmidrule(lr){2-10}
& Rural/Urban & Count & Reasoning & Status
& Category & Shape & Color & Background & Avg. \\
\midrule
\rowcolor[gray]{0.94}
\multicolumn{10}{@{}l}{\textit{Commercial MLLMs}} \\
GPT-4o~\cite{openai2024gpt4o} & 55.19 & 16.50 & 21.39 & 19.22 & 16.26 & 39.32 & 47.45 & 18.78 & 29.26 \\
Gemini-2.5-Flash~\cite{geminiteam2025} & 56.49 & \BestScore{18.49} & 20.40 & 18.69 & 14.81 & 34.17 & 36.08 & 16.11 & 26.91 \\
\midrule
\rowcolor[gray]{0.94}
\multicolumn{10}{@{}l}{\textit{General-purpose MLLMs}} \\
Qwen2.5-VL-7B~\cite{bai2025qwen25vl} & 47.12 & 16.18 & 19.50 & 9.20 & 16.31 & 21.81 & 51.76 & 17.96 & 24.98 \\
Qwen3-VL-8B~\cite{bai2025qwen3vl} & 56.47 & \SecondScore{18.35} & 23.40 & \SecondScore{21.20} & 19.76 & 41.24 & 56.47 & 20.00 & 32.11 \\
LLaVA-v1.6-7B~\cite{liu2024llavanext} & 52.00 & 13.68 & 19.80 & 17.40 & 15.91 & 29.77 & 41.31 & 17.96 & 25.98 \\
LLaVA-OV-7B~\cite{li2024llavaonevision} & 50.08 & 11.68 & 21.80 & 11.20 & 19.15 & 31.98 & 45.49 & 17.14 & 26.07 \\
\midrule
\rowcolor[gray]{0.94}
\multicolumn{10}{@{}l}{\textit{Remote sensing MLLMs}} \\
% RSUniVLM & \TBD & \TBD & \TBD & \TBD & \TBD & \TBD & \TBD & \TBD & \TBD \\
GeoChat~\cite{kuckreja2024geochat} & \SecondScore{61.42} & 11.76 & 16.70 & 6.50 & 8.00 & 21.47 & 17.39 & 11.84 & 19.39 \\
VHM~\cite{pang2025vhm} & 56.39 & 12.26 & 18.80 & 13.40 & 17.12 & 31.86 & 46.27 & 12.24 & 26.04 \\
GeoLLaVA-8K~\cite{geollava8k2025} & 54.68 & 12.83 & 21.32 & 4.41 & 14.81 & 22.41 & 49.52 & 16.18 & 24.52 \\
GeoLens~\cite{geolens2026} & 51.52 & 15.35 & 22.40 & 10.70 & 15.60 & 43.28 & 50.98 & 16.33 & 28.27 \\
ZoomEarth~\cite{liu2026zoomearth} & 39.70 & 9.34 & 15.50 & 9.50 & 11.04 & 37.40 & 29.67 & 6.94 & 19.89 \\
% ImageRAG & 58.55 & 13.70 & 21.20 & 10.00 & 21.07 & 33.90 & 45.75 & 19.18 & 27.92 \\
% ZoomEye & 48.24 & 14.76 & 22.10 & 10.40 & 23.61 & 28.58 & 45.36 & 21.63 & 26.84 \\
% RAP & 45.45 & 16.78 & 26.10 & 11.00 & 24.32 & 30.96 & 51.90 & 24.08 & 28.82 \\
\midrule
\rowcolor[gray]{0.94}
\multicolumn{10}{@{}l}{\textit{Training-free evidence acquisition methods}} \\
ZoomSearch~\cite{zoomsearch2025} & 56.63 & 13.59 & \SecondScore{25.50} & 18.60 & 24.21 & 45.54 & 52.42 & \BestScore{24.49} & \SecondScore{32.62} \\
WeaveEarth~\cite{weaveearth2026} & \BestScore{63.66} & 12.43 & 19.20 & 16.80 & 17.73 & \BestScore{50.62} & \BestScore{57.52} & 10.20 & 31.02 \\
% \midrule
% \multicolumn{10}{@{}l}{\textit{General focus methods for MLLMs}} \\
% AttWarp & \TBD & \TBD & \TBD & \TBD & \TBD & \TBD & \TBD & \TBD & \TBD \\
% ViCrop & \TBD & \TBD & \TBD & \TBD & \TBD & \TBD & \TBD & \TBD & \TBD \\
\midrule
\rowcolor[gray]{0.94}
\multicolumn{10}{@{}l}{\textit{Ours}} \\
\textbf{GeoJeV (Qwen3-VL-8B)} & \SecondScore{61.42} & 15.18 & \BestScore{26.80} & 19.10 & \BestScore{26.14} & \SecondScore{48.81} & \SecondScore{57.12} & 20.00 & \BestScore{34.32} \\
\textbf{GeoJeV (LLaVA-OV-7B)} & 55.91 & 15.18 & 12.80 & \BestScore{26.60} & \SecondScore{25.84} & 37.51 & 51.11 & \SecondScore{20.41} & 30.67 \\
\bottomrule
\end{tabular*}
\par\vspace{2pt}
\begin{tablenotes}[flushleft]
\footnotesize\raggedright
\item[] \textbf{Avg.}: Unweighted mean across categories. Best and second-best results are in \textbf{bold} and \underline{underlined}, respectively.
\end{tablenotes}
\end{threeparttable}
\end{table*}
% ===== End inlined source: tables/main_lrsvqa.tex =====

% ===== Begin inlined source: tables/efficiency.tex =====
\begin{table*}[!tp]
\centering
\caption{Accuracy--latency comparison across three benchmarks with Qwen3-VL-8B.}
\label{tab:efficiency}

\small
\setlength{\tabcolsep}{4pt}
\renewcommand{\arraystretch}{1.10}

\begin{threeparttable}
\begin{tabular*}{\textwidth}{
@{\extracolsep{\fill}}
l
cc
cc
cc
cc
@{}
}
\toprule

\multirow{2}{*}{\textbf{Method}}
& \multicolumn{2}{c}{\textbf{LRS-GRO}}
& \multicolumn{2}{c}{\textbf{MME-RealWorld-RS}}
& \multicolumn{2}{c}{\textbf{LRS-VQA}}
& \multicolumn{2}{c}{\textbf{Avg.}} \\

\cmidrule(lr){2-3}
\cmidrule(lr){4-5}
\cmidrule(lr){6-7}
\cmidrule(lr){8-9}

& Acc. (\%)$\uparrow$ & Lat. (s/Q)$\downarrow$
& Acc. (\%)$\uparrow$ & Lat. (s/Q)$\downarrow$
& Acc. (\%)$\uparrow$ & Lat. (s/Q)$\downarrow$
& Acc. (\%)$\uparrow$ & Lat. (s/Q)$\downarrow$ \\

\midrule

Direct
& 57.68 & \BestScore{7.97}
& 41.94 & \SecondScore{14.70}
& 32.11 & \BestScore{8.74}
& 43.91 & \SecondScore{10.47} \\

ZoomSearch
& 55.32 & 69.62
& 38.17 & 122.08
& \SecondScore{32.62} & 78.49
& 42.04 & 90.06 \\

WeaveEarth
& \SecondScore{59.86} & \SecondScore{8.06}
& \SecondScore{47.21} & \BestScore{13.27}
& 31.02 & \SecondScore{8.95}
& \SecondScore{46.03} & \BestScore{10.09} \\

\midrule

\textbf{GeoJeV}
& \BestScore{60.59} & 8.52
& \BestScore{52.79} & 15.88
& \BestScore{34.32} & 10.37
& \BestScore{49.23} & 11.59 \\

\bottomrule
\end{tabular*}

\par\vspace{2pt}
\begin{tablenotes}[flushleft]
\footnotesize\raggedright
\item[] \textbf{Acc.}: Accuracy (\%). \textbf{Lat.}: End-to-end inference latency in seconds per question (s/Q). \textbf{Avg.}: Unweighted mean across benchmarks.
Best and second-best results are in \textbf{bold} and \underline{underlined}, respectively.
\end{tablenotes}
\end{threeparttable}
\end{table*}

% ===== End inlined source: tables/efficiency.tex =====

\begin{table}[!t]
\centering
\caption{Accuracy comparison for cross-backbone generalization of GeoJeV on three benchmarks.}
\label{tab:backbone_ablation}
\small
\setlength{\tabcolsep}{3pt}
\renewcommand{\arraystretch}{1.10}
\begin{threeparttable}
\begin{tabularx}{\columnwidth}{@{}l*{4}{>{\centering\arraybackslash}X}@{}}
\toprule
\textbf{Method}
& \textbf{LRS-GRO}
& \textbf{MME-RS}
& \textbf{LRS-VQA}
& \textbf{Avg.} \\
\midrule

\rowcolor[gray]{0.94}
\multicolumn{5}{@{}l@{}}{\textit{Qwen3-VL-8B}} \\
Direct
& \SecondScore{57.68} & \SecondScore{41.94} & \SecondScore{32.11} & \SecondScore{43.91} \\
\textbf{GeoJeV}
& \BestScore{60.59} & \BestScore{52.79}
& \BestScore{34.32} & \BestScore{49.23} \\
\textit{Gain}
& $+2.91\pp$ & $+10.85\pp$
& $+2.21\pp$ & $+5.32\pp$ \\

\midrule

\rowcolor[gray]{0.94}
\multicolumn{5}{@{}l@{}}{\textit{LLaVA-OV-7B}} \\
Direct
& \SecondScore{46.18} & \SecondScore{26.84} & \SecondScore{26.07} & \SecondScore{33.03} \\
\textbf{GeoJeV}
& \BestScore{55.40} & \BestScore{55.88}
& \BestScore{30.67} & \BestScore{47.32} \\
\textit{Gain}
& $+9.22\pp$ & $+29.04\pp$
& $+4.60\pp$ & $+14.29\pp$ \\

\bottomrule
\end{tabularx}
\par\vspace{2pt}
\begin{tablenotes}[flushleft]
\footnotesize\raggedright
\item[]
\textbf{Avg.}: Unweighted mean across benchmarks.
Best and second-best accuracies within each backbone are in \textbf{bold} and \underline{underlined}, respectively.
\end{tablenotes}
\end{threeparttable}
\end{table}
% ===== End inlined source: tables/backbone_ablation.tex =====

\noindent \textbf{Implementation Details.}
For search control, we use \texttt{metask-jev-4b-policy-\allowbreak mix}\footnote{\url{https://huggingface.co/wayfind/metask-jev-4b-policy-mix}} as the JeV model. Qwen3-VL-8B is the default answering MLLM, and LLaVA-OV-7B is an additional answering backbone for cross-backbone evaluation. The JeV model and each answering MLLM remain frozen throughout inference.

For evidence acquisition, we use a $3\times3$ grid with temperature $T=2.0$. The shared cap $K=6$ limits both the number of retained children per expansion and the final number of evidence regions. Candidate regions must be at least $\delta=224$ pixels in both width and height. We set a common budget of $M=10$ JeV forward passes per question across all benchmarks. Together, these limits help avoid excessive exploration of small regions that may lack sufficient spatial context while keeping computational overhead manageable. Depth limits and early stopping can further reduce the number of forward passes.

The maximum search depth depends on the image resolution
and minimum crop size:
\[
    L=1+\left\lfloor\log_g\frac{\min(W,H)}{\delta}\right\rfloor,
\]
where $W$ and $H$ are the image width and height. We set $L$ per benchmark from its image resolution and hold it
fixed for all images. We enable
Depth Gate at all intermediate depths,
with $\mathcal D_{\mathrm{gate}}=\{2,\ldots,L-1\}$.

For each JeV forward pass, the image processor dynamically adjusts the input resolution while preserving the aspect ratio, with a pixel budget ranging from $256^2$ to $4096^2$. The selected regions are combined into a spatial evidence canvas according to their locations, without an explicit limit on canvas size. The answering MLLM then processes the canvas using dynamic resolution for Qwen3-VL-8B or any-resolution tiling for LLaVA-OV-7B. Final answering uses greedy decoding with a maximum of 1,024 new tokens.

\noindent \textbf{Inference Environment.}
Local inference is performed on an NVIDIA GeForce RTX 5090 GPU (32 GB) with Ubuntu 22.04, Python 3.12, PyTorch 2.12.1, and CUDA 13.0. JeV probability normalization is performed in float32.

% Position experiment float pages without changing fonts or table styles.
\makeatletter
\setlength{\@dblfptop}{0pt}
\setlength{\@dblfpsep}{3pt}
\setlength{\@dblfpbot}{0pt plus 1fil}
\makeatother

\subsection{Overall and Category-wise Performance}
\label{sec:experiment_main_results}

We evaluate the overall and category-wise performance of GeoJeV on three UHR remote sensing benchmarks. Tables~\ref{tab:main_lrs_mme} and~\ref{tab:main_lrsvqa} present comparisons with the baseline methods introduced in Section~\ref{sec:experiment_setup}.

\textit{i) Results on LRS-GRO.}
GeoJeV with Qwen3-VL-8B reaches an average accuracy of 60.59\%, between the training-based GeoLens at 61.83\% and the training-free WeaveEarth at 59.86\%. Relative to direct answering, the Region and Object categories improve by $+9.02$ and $+2.92\pp$, while Global drops by $3.23\pp$. Focusing on task-relevant regions helps the model identify small objects and local structures, but may limit its understanding of the overall scene.

\textit{ii) Results on MME-RealWorld-RS.}
GeoJeV improves the average accuracy of Qwen3-VL-8B from 41.94\% to 52.79\% with a gain of $+10.85\pp$, and outperforms both training-free baselines. With LLaVA-OV-7B, it achieves the highest average accuracy of 55.88\% among all compared methods. However, GeoJeV with Qwen3-VL-8B achieves 29.93\% in the Count category, below the 33.28\% achieved by Gemini-2.5-Flash. This suggests that selecting only a few regions may miss some objects spread across the image for accurate counting.

\textit{iii) Results on LRS-VQA.}
GeoJeV with Qwen3-VL-8B achieves the highest average accuracy of 34.32\%, exceeding ZoomSearch and WeaveEarth by $+1.70$ and $+3.30\pp$. It also achieves the highest accuracy in Reasoning and Category among all compared methods. Relative to direct answering, Count and Status decrease by $3.17$ and $2.10\pp$. With LLaVA-OV-7B, GeoJeV achieves the highest Status accuracy but loses $9.00\pp$ on Reasoning compared with its own direct baseline. These results suggest that the benefits of selective evidence acquisition vary across question categories and answering backbones.

\begin{figure}[!t]
\centering
\includegraphics[width=0.9\columnwidth]{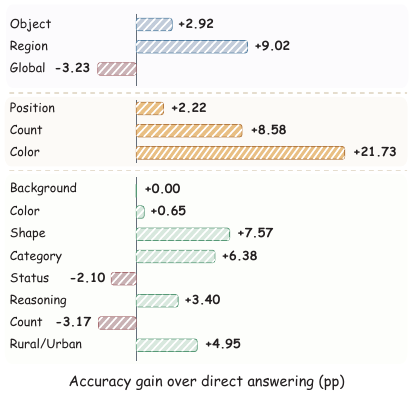}
\caption{Category-wise accuracy gains of GeoJeV over direct answering with Qwen3-VL-8B on three benchmarks.}
\label{fig:gains}
\end{figure}

\begin{figure}[!t]
\centering
\includegraphics[width=0.95\columnwidth]{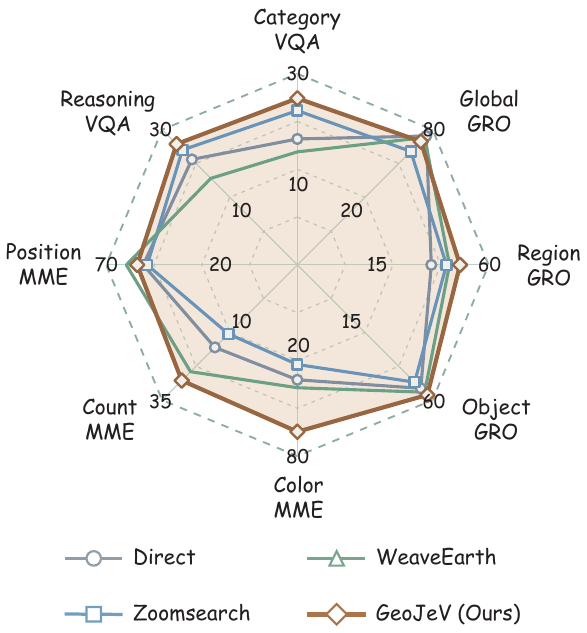}
\caption{Accuracy comparison of four methods across eight categories from three benchmarks using Qwen3-VL-8B.}
\label{fig:radar_lsvqa}
\end{figure}

Fig.~\ref{fig:gains} shows the accuracy gains over direct answering across all 14 categories. The Color category on MME-RealWorld-RS has the largest gain of $+21.73\pp$. Most of the remaining gains above $+6\pp$ are in Region, Category, and Shape. The Count category on MME-RealWorld-RS also improves by $+8.58\pp$ from a baseline of 21.35\%. However, its accuracy remains the lowest among the three MME-RealWorld-RS categories. Compared with direct answering, GeoJeV shows lower accuracy in the Count and Status categories of LRS-VQA and the Global category of LRS-GRO. Overall, the larger gains are mainly found in categories that depend on local visual evidence.

Fig.~\ref{fig:radar_lsvqa} compares four methods across eight categories from the three benchmarks. GeoJeV achieves the highest accuracy in six of these categories. Its largest advantage is in the Color category on MME-RealWorld-RS, with 69.80\% accuracy compared with 51.39\% for WeaveEarth, the strongest baseline shown in this category. In the Global category on LRS-GRO, the top three methods differ by less than $3.3\pp$. This may indicate that these questions depend more on global scene information than on local visual details.

Overall, GeoJeV improves average accuracy across the three benchmarks, with larger gains in categories that depend on local visual evidence. Questions requiring broad spatial coverage remain challenging, suggesting that better coverage of the overall scene is still needed.

% ===== Begin inlined source: tables/component_ablation.tex =====

\begin{table*}[!tp]
\centering
\caption{Component ablations of GeoJeV on MME-RealWorld-RS with Qwen3-VL-8B.}
\label{tab:ablation}
\small
\setlength{\tabcolsep}{3pt}
\renewcommand{\arraystretch}{1.10}

\begin{threeparttable}
\begin{tabular*}{\textwidth}{
@{\extracolsep{\fill}}
l
ccc
cccc
cc
@{}
}
\toprule
\multirow{2}{*}{\textbf{Variant}}
& \multicolumn{3}{c}{\textbf{Components}}
& \multicolumn{4}{c}{\textbf{MME-RealWorld-RS (\%) $\uparrow$}}
& \multicolumn{2}{c}{\textbf{Inference Cost}} \\
\cmidrule(lr){2-4}
\cmidrule(lr){5-8}
\cmidrule(lr){9-10}

& \makecell{Region\\Rank}
& \makecell{Sparse\\Branch}
& \makecell{Depth\\Gate}
& Color & Count & Position & Avg.
& \makecell{Fwd./Q\\$\downarrow$}
& \makecell{Lat. (s/Q)\\$\downarrow$} \\
\midrule

Direct
& -- & -- & --
& 48.07 & 21.35 & 56.41 & 41.94
& \BestScore{0.00} & \BestScore{14.70} \\

+ Region Rank
& \cmark & -- & --
& \SecondScore{67.17} & 28.06 & \SecondScore{60.06} & 51.76
& 2.91 & 17.17 \\

+ Sparse Branch
& \cmark & \cmark & --
& 66.53 & \SecondScore{28.14} & \BestScore{60.86} & \SecondScore{51.84}
& 2.60 & 16.31 \\

\midrule
\textbf{GeoJeV (Full)}
& \cmark & \cmark & \cmark
& \BestScore{69.80} & \BestScore{29.93}
& 58.63 & \BestScore{52.79}
& \SecondScore{2.38} & \SecondScore{15.88} \\

\bottomrule
\end{tabular*}

\par\vspace{2pt}
\begin{tablenotes}[flushleft]
\footnotesize\raggedright
\item[]
\textbf{\cmark}: Component enabled. \textbf{--}: Disabled or not applicable.
\textbf{Avg.}: Unweighted mean of category accuracies.
\textbf{Fwd./Q}: Mean JeV forward passes per question.
\textbf{Lat.}: End-to-end inference latency in seconds per question (s/Q).
Best and second-best results are in \textbf{bold} and \underline{underlined}, respectively.
\end{tablenotes}
\end{threeparttable}
\end{table*}

% ===== End inlined source: tables/component_ablation.tex =====

\subsection{Qualitative Analysis}
\label{sec:experiment_qualitative}

\begin{figure*}[t]
\centering
\includegraphics[width=0.95\textwidth]{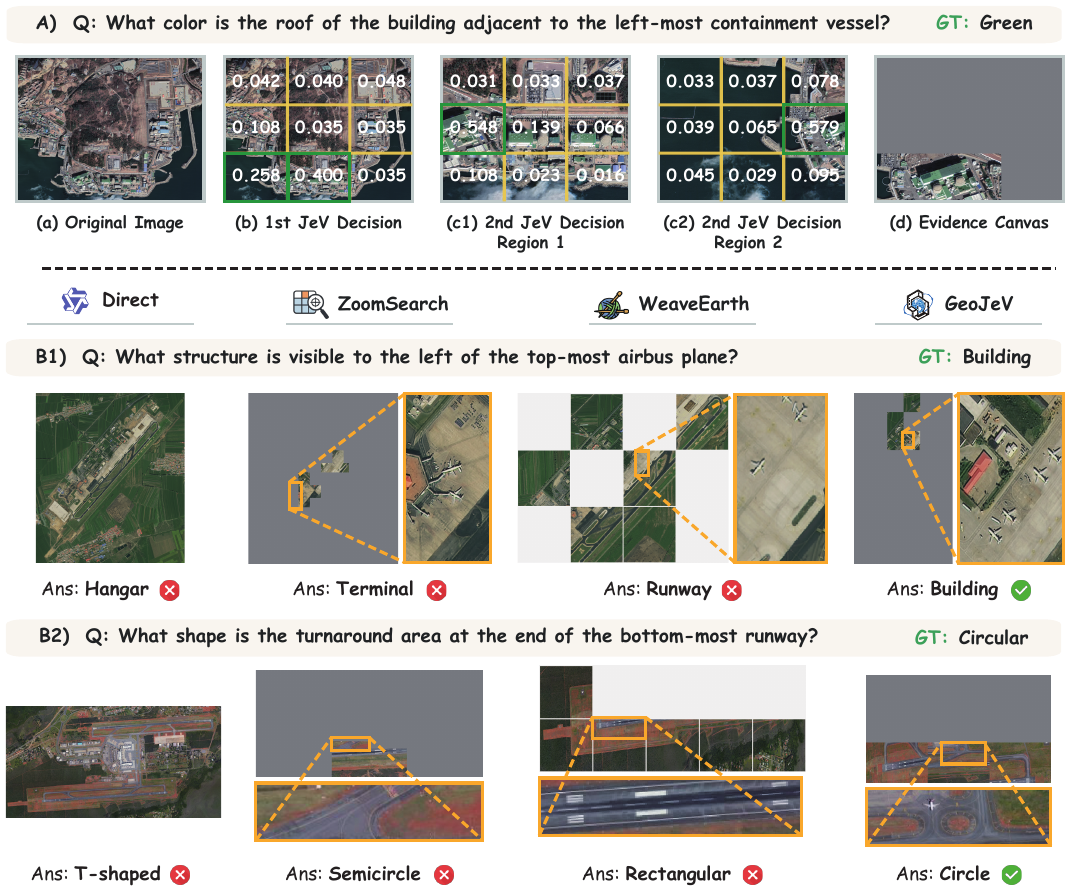}
\caption{Qualitative results on LRS-VQA. A) The complete evidence acquisition process of GeoJeV. B) Evidence images of Direct, ZoomSearch, WeaveEarth, and GeoJeV on two questions. Green and red markers indicate correct and incorrect answers.}
\label{fig:qualitative}
\end{figure*}

Fig.~\ref{fig:qualitative} presents qualitative examples from one question.
In Fig.~\ref{fig:qualitative}(a), the spatial probability distributions
estimated by JeV guide the search toward regions relevant to the question.
The selected crops preserve useful local details and are combined into
a spatial evidence canvas for the answering MLLM.
Fig.~\ref{fig:qualitative}(b) compares the visual inputs of Direct,
ZoomSearch, WeaveEarth, and GeoJeV on two additional questions.
In both examples, GeoJeV preserves the relevant objects and surrounding
details needed for answering. These examples suggest that probability-guided
search can provide clearer visual evidence and support more accurate answering.

\setcounter{dbltopnumber}{1}

\subsection{Efficiency and Generalization}
\label{sec:experiment_efficiency_generalization}

\subsubsection{Accuracy--Latency Trade-off}
\label{sec:experiment_efficiency}

\begin{table}[!t]
\centering
\caption{Sensitivity analysis of grid size $g$ and softmax temperature $T$ on MME-RealWorld-RS with Qwen3-VL-8B.}
\label{tab:grid_temperature}
\small
\setlength{\tabcolsep}{2pt}
\renewcommand{\arraystretch}{1.10}
\begin{threeparttable}
\begin{tabular*}{\columnwidth}{@{\extracolsep{\fill}}l*{7}{c}@{}}
\toprule
\textbf{Setting}
& \textbf{Default}
& Color & Count & Position & Avg.
& \makecell{Fwd./Q\\$\downarrow$}
& \makecell{Lat. (s/Q)\\$\downarrow$} \\
\midrule
\rowcolor[gray]{0.94}
\multicolumn{8}{@{}l}{\textit{(a) Grid size $g$}} \\
$g=2$ & -- & \SecondScore{68.05} & \BestScore{32.14} & 55.29 & \SecondScore{51.83} & \BestScore{1.99} & 22.03 \\
$g=3$ & \cmark & \BestScore{69.80} & \SecondScore{29.93} & \BestScore{58.63} & \BestScore{52.79} & \SecondScore{2.38} & \SecondScore{15.88} \\
$g=4$ & -- & 62.23 & 22.35 & \SecondScore{56.40} & 46.99 & 2.50 & \BestScore{15.74} \\
\midrule
\rowcolor[gray]{0.94}
\multicolumn{8}{@{}l}{\textit{(b) Temperature $T$}} \\
$T=1.5$ & -- & \SecondScore{64.30} & \SecondScore{27.98} & 52.74 & 48.34 & \BestScore{1.70} & 19.64 \\
$T=2.0$ & \cmark & \BestScore{69.80} & \BestScore{29.93} & \BestScore{58.63} & \BestScore{52.79} & \SecondScore{2.38} & \BestScore{15.88} \\
$T=3.0$ & -- & 64.10 & 27.20 & \SecondScore{54.80} & \SecondScore{48.70} & 2.71 & \SecondScore{17.43} \\
\bottomrule
\end{tabular*}
\par\vspace{2pt}
\begin{tablenotes}[flushleft]
\footnotesize\raggedright
\item[]
\textbf{\cmark}: Default settings.
\textbf{Avg.}: Unweighted mean of category accuracies.
\textbf{Fwd./Q}: Mean JeV forward passes per question.
\textbf{Lat.}: End-to-end inference latency in seconds per question (s/Q).
Best and second-best results within each panel are in \textbf{bold} and \underline{underlined}, respectively.
\end{tablenotes}
\end{threeparttable}
\end{table}

\begin{figure*}[t]
\centering
\includegraphics[width=0.95\textwidth]{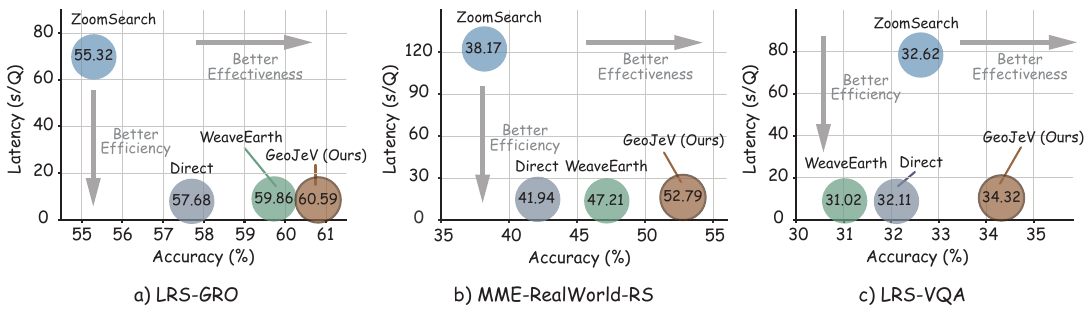}
\caption{Accuracy--latency comparison of Direct, ZoomSearch, WeaveEarth, and GeoJeV across three benchmarks with Qwen3-VL-8B. GeoJeV achieves the highest accuracy with a small increase in latency over Direct.}
\label{fig:acc_lat}
\end{figure*}

Table~\ref{tab:efficiency} and Fig.~\ref{fig:acc_lat} compare four methods using the same frozen Qwen3-VL-8B answering backbone. GeoJeV achieves the highest average accuracy of 49.23\%. Compared with direct answering, it improves accuracy by $+5.32\pp$ with only 1.12 additional seconds per question. It also outperforms ZoomSearch by $7.19\pp$ and achieves a $7.8\times$ speedup. WeaveEarth is 1.50 seconds faster per question than GeoJeV, but its accuracy is $3.20\pp$ lower. Overall, GeoJeV achieves a favorable accuracy--latency trade-off through evidence acquisition guided by spatial probability distributions, without repeated decoding of intermediate decisions.

\subsubsection{Cross-Backbone Generalization}
\label{sec:experiment_backbones}

To evaluate generalization across answering backbones, we use Qwen3-VL-8B and LLaVA-OV-7B with the same frozen JeV model and search settings, without task-specific parameter updates. Table~\ref{tab:backbone_ablation} reports the accuracy gains over direct answering. GeoJeV improves average accuracy on all three benchmarks with both backbones. LLaVA-OV-7B achieves an average gain of $+14.29\pp$ and a maximum gain of $+29.04\pp$ on MME-RealWorld-RS. These results support the use of the same evidence acquisition process across answering backbones. The model with lower direct-answering accuracy benefits more from explicit evidence selection.

\setcounter{dbltopnumber}{2}

\subsection{Component Ablation}
\label{sec:experiment_components}

As summarized in Table~\ref{tab:ablation}, we evaluate the contribution of each component on MME-RealWorld-RS using Qwen3-VL-8B. Direct serves as the baseline without evidence acquisition. The first search variant uses Region Rank with fixed Top-2 branching and no early stopping. Sparse Branch is then enabled to adjust the branching width. Finally, Depth Gate adds early stopping to form the full GeoJeV method.

\noindent\textbf{Region Rank.}
With fixed Top-2 branching, Region Rank improves average accuracy by $+9.82\pp$ over direct answering, with gains in all three categories. These results show that ranking regions by spatial probabilities helps select task-relevant visual evidence and improve answering accuracy.

\noindent\textbf{Sparse Branch.}
With Sparse Branch, average accuracy increases slightly by $+0.08\pp$. Inference latency decreases by $5.0\%$ from 17.17 to 16.31 seconds per question. By adjusting the branching width to the distribution's sparse support, Sparse Branch can avoid unnecessary region evaluations. Overall, these results show that adaptive branching improves search efficiency while preserving answering accuracy.

\noindent\textbf{Depth Gate.}
With Depth Gate, average accuracy increases by a further $+0.95\pp$. JeV forward passes decrease to 2.38 and latency falls to 15.88 seconds per question. Both are the lowest among the search variants. However, Position accuracy decreases from 60.86\% to 58.63\%. Overall, Depth Gate improves average accuracy and reduces search cost, although some Position questions may benefit from deeper exploration.

\subsection{Sensitivity Analysis}
\label{sec:experiment_sensitivity}

This section examines how branching width, grid size, and temperature affect answering accuracy and evidence acquisition cost on MME-RealWorld-RS with Qwen3-VL-8B.

\begin{figure}[!t]
\centering
\includegraphics[width=0.9\columnwidth]{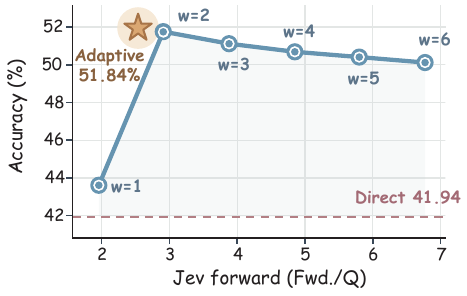}
\caption{Accuracy and JeV forward passes for fixed widths and adaptive branching using Sparse Branch, both with Depth Gate disabled. The dashed line marks direct answering.}
\label{fig:width_curve}
\end{figure}

\noindent\textbf{Branch Width.}
Fig.~\ref{fig:width_curve} compares fixed widths $w\in\{1,\ldots,6\}$ with the adaptive width of Sparse Branch, with Depth Gate disabled in all cases. With fixed branching, accuracy rises from 43.61\% at $w=1$ to a peak of 51.76\% at $w=2$. Increasing the width further requires more JeV forward passes but reduces accuracy to 50.12\% at $w=6$. In comparison, Sparse Branch achieves the highest accuracy of 51.84\% and uses 10.7\% fewer forward passes than the best fixed width. A single branch may miss useful evidence after an incorrect initial selection, and larger fixed widths may spend computation on less useful regions. Overall, adapting the width to region probabilities provides a better balance between accuracy and search cost than the tested fixed widths.

\noindent\textbf{Grid Granularity.}
Table~\ref{tab:grid_temperature}(a) compares grid sizes with the maximum depth fixed at $L=3$. Relative to the default $3\times3$ grid, the $2\times2$ grid reduces average accuracy by $0.96\pp$ and has the highest latency of 22.03 seconds per question. Its larger crops may increase the cost of final answering. The $4\times4$ grid reduces average accuracy by $5.80\pp$, with the largest losses in Color and Count. This suggests that finer grids may split objects across crops and reduce useful context. Overall, the default $3\times3$ grid balances local detail and the spatial context needed for answering.

\noindent\textbf{Temperature.}
Table~\ref{tab:grid_temperature}(b) compares softmax temperatures in Eq.~\eqref{eq:spatial_probability}. Relative to the default $T=2.0$, accuracy decreases by $4.45\pp$ at $T=1.5$ and $4.09\pp$ at $T=3.0$. Latency also increases to 19.64 and 17.43 seconds per question, respectively. A lower temperature produces a sharper distribution and narrows the sparse support, which may cause the search to stop too early. A higher temperature makes candidate probabilities more similar. These changes may also affect the size of the evidence canvas and the cost of final answering. Among the tested temperatures, the default $T=2.0$ provides the highest accuracy and the lowest latency.

Branching width, grid size, and temperature affect how the spatial probability distribution is formed and used. Sparse Branch and Depth Gate help reduce unnecessary exploration. However, using fewer grid cells or stopping the search too early may leave larger crops in the evidence canvas and increase the cost of final answering. Fewer JeV forward passes therefore do not always lead to lower latency. These results highlight the need to consider both search computation and the cost of processing the final evidence canvas.

% ===== End inlined source: sec/4_experiment.tex =====

% ===== Begin inlined source: sec/5_conclusion.tex =====
\section{Conclusion}
\label{sec:conclusion}

We presented GeoJeV, a training-free method for efficient UHR
remote sensing understanding that formulates evidence acquisition
as spatial probability-driven decision making. A frozen JeV model
estimates a spatial probability distribution over candidate regions
at each evaluated node. Region Rank, Sparse Branch, and Depth Gate
use this shared distribution to control search priority, branching
width, and refinement depth, respectively. This formulation moves
search control beyond RLHF-aligned generation and
eliminates repeated autoregressive decoding of intermediate decisions.
The acquired regions are combined into a spatial evidence canvas
for final answering by a separate frozen MLLM. Experiments on three
UHR remote sensing benchmarks show an average gain of $+5.32\pp$
over direct answering at $1.12$ additional seconds per question and
a $7.8\times$ speedup over the recent method.

The benefits of selective evidence acquisition remain uneven across
question categories, with some category accuracies below those of direct
answering. Future work will examine the evidence requirements of
different questions to better balance local visual details, spatial
coverage, and global scene context. In particular, we will explore
region selection with broader spatial coverage and improved spatial
evidence construction for questions involving distributed objects.
These directions aim to improve consistency across question
categories while retaining efficient and training-free evidence acquisition.

% ===== End inlined source: sec/5_conclusion.tex =====

\bibliographystyle{IEEEtran}
\bibliography{GeoJeV}

\vfill

\end{document}